\documentclass{article}

\usepackage{microtype}
\usepackage{graphicx}
\usepackage{subfigure}
\usepackage{booktabs} % for professional tables

\usepackage{hyperref}
\usepackage{multirow}
\usepackage{makecell}
\usepackage{listings}
\usepackage[accepted]{icml2024}

\usepackage{amsmath}
\usepackage{amssymb}
\usepackage{mathtools}
\usepackage{amsthm}

\usepackage[capitalize,noabbrev]{cleveref}

\usepackage{colortbl} % 包含colortbl包

\theoremstyle{plain}

\theoremstyle{definition}

\theoremstyle{remark}

\usepackage[textsize=tiny]{todonotes}
\usepackage{CJKutf8} % for Chinese
\usepackage{listings} % pseudocode
\usepackage{xcolor}
\usepackage{tcolorbox} % table
\usepackage{amsmath}
\DeclareMathOperator*{\argmax}{arg\,max}

\definecolor{codegreen}{rgb}{0,0.6,0}
\definecolor{codegray}{rgb}{0.5,0.5,0.5}
\definecolor{codepurple}{rgb}{0.58,0,0.82}
\definecolor{backcolour}{rgb}{0.95,0.95,0.92}

\lstdefinestyle{mystyle}{
    backgroundcolor=\color{backcolour},   
    commentstyle=\color{codegreen},
    keywordstyle=\color{magenta},
    numberstyle=\tiny\color{codegray},
    stringstyle=\color{codepurple},
    basicstyle=\ttfamily\footnotesize,
    breakatwhitespace=false,         
    breaklines=true,                 
    captionpos=b,                    
    keepspaces=true,                 
    numbers=left,                    
    numbersep=5pt,                  
    showspaces=false,                
    showstringspaces=false,
    showtabs=false,                  
    tabsize=2
}

\icmltitlerunning{LLM-Detector: Improving AI-Generated Chinese Text Detection with Open-Source LLM Instruction Tuning}

\begin{document}
\begin{CJK*}{UTF8}{gbsn}
\twocolumn[
\icmltitle{LLM-Detector: Improving AI-Generated Chinese Text Detection with \\ Open-Source LLM Instruction Tuning}



\icmlsetsymbol{equal}{*}

\begin{icmlauthorlist}
\icmlauthor{Rongsheng Wang}{mpu}
\icmlauthor{Haoming Chen}{mpu}
\icmlauthor{Ruizhe Zhou}{mpu}
\icmlauthor{Han Ma}{mpu}
\icmlauthor{Yaofei Duan}{mpu}
\icmlauthor{Yanlan Kang}{fudan}
\icmlauthor{Songhua Yang}{wuhan}
\icmlauthor{Baoyu Fan}{mpu}
\icmlauthor{Tao Tan}{mpu}
\end{icmlauthorlist}

\icmlaffiliation{mpu}{Macao Polytechnic University}
\icmlaffiliation{fudan}{Fudan University}
\icmlaffiliation{wuhan}{Wuhan University}

\icmlcorrespondingauthor{Tao Tan}{taotan@mpu.edu.mo}

\icmlkeywords{Machine Learning, ICML}

\vskip 0.3in
]



\printAffiliationsAndNotice{}  % leave blank if no need to mention equal contribution

%%%%%%%%%%%%%%%%%%%%%%%%%%%%%%%%%%%%%%%%%%%%%%%%%%%%%%%%%%%%%%%%%%%%%%%%%%%%%%%
%%%%%%%%%%%%%%%%%%%%%%%%%%%%%%%%%%%%%%%%%%%%%%%%%%%%%%%%%%%%%%%%%%%%%%%%%%%%%%%
% Abstract
% Check By Rongsheng Wang, 
%%%%%%%%%%%%%%%%%%%%%%%%%%%%%%%%%%%%%%%%%%%%%%%%%%%%%%%%%%%%%%%%%%%%%%%%%%%%%%%
%%%%%%%%%%%%%%%%%%%%%%%%%%%%%%%%%%%%%%%%%%%%%%%%%%%%%%%%%%%%%%%%%%%%%%%%%%%%%%%
\begin{abstract}
ChatGPT and other general large language models (LLMs) have achieved remarkable success, but they have also raised concerns about the misuse of AI-generated texts. Existing AI-generated text detection models, such as based on BERT and RoBERTa, are prone to in-domain over-fitting, leading to poor out-of-domain (OOD) detection performance. In this paper, we first collected Chinese text responses generated by human experts and 9 types of LLMs, for which to multiple domains questions, and further created a dataset that mixed human-written sentences and sentences polished by LLMs. We then proposed LLM-Detector, a novel method for both document-level and sentence-level text detection through Instruction Tuning of LLMs. Our method leverages the wealth of knowledge LLMs acquire during pre-training, enabling them to detect the text they generate. Instruction tuning aligns the model's responses with the user's expected text detection tasks. Experimental results show that previous methods struggle with sentence-level AI-generated text detection and OOD detection. In contrast, our proposed method not only significantly outperforms baseline methods in both sentence-level and document-level text detection but also demonstrates strong generalization capabilities. Furthermore, since LLM-Detector is trained based on open-source LLMs, it is easy to customize for deployment.
\end{abstract}

%%%%%%%%%%%%%%%%%%%%%%%%%%%%%%%%%%%%%%%%%%%%%%%%%%%%%%%%%%%%%%%%%%%%%%%%%%%%%%%
%%%%%%%%%%%%%%%%%%%%%%%%%%%%%%%%%%%%%%%%%%%%%%%%%%%%%%%%%%%%%%%%%%%%%%%%%%%%%%%
% Introduction
% Check by Rongsheng Wang
%%%%%%%%%%%%%%%%%%%%%%%%%%%%%%%%%%%%%%%%%%%%%%%%%%%%%%%%%%%%%%%%%%%%%%%%%%%%%%%
%%%%%%%%%%%%%%%%%%%%%%%%%%%%%%%%%%%%%%%%%%%%%%%%%%%%%%%%%%%%%%%%%%%%%%%%%%%%%%%
\section{Introduction}
\label{introduction}
Large language models (LLMs), such as ChatGPT, represent a significant milestone in the field of natural language processing (NLP).
LLMs have been pre-trained on extensive text corpora, enabling them to generate texts that are contextually relevant and fluent.~\cite{brown2020language}
However, the impressive capabilities of generative language models in text generation have also led to rising concerns about their possible misuse in various areas, including phishing, dissemination of misinformation, and academic dishonesty.
% Additionally, with the application of LLMs products like ChatGPT, the anticipated increase in AI-generated text data in the future of AI-generated text data has the potential to contaminate genuine human generated data~\cite{hataya2023will}.
Unfortunately, when human classifying AI-generated and human-written texts, human only slightly better than random guessing~\cite{gehrmann2019gltr}.
Therefore, our goal is to develop an automated system to classify AI-generated texts with the aim of mitigating its potential for misuse.

There has been some previous effort in detecting AI-generated texts.
First, Guo et al.~\cite{guo2023close} fine-tuned RoBERTa to detect whether a certain text (English and Chinese) is generated by ChatGPT or written by human.
However, the study conducted demonstrates that a limitation of supervised models is the potential occurrence of overfitting in-domain data, resulting in poor detection performance out-of-domain (OOD)~\cite{chakraborty2023possibilities}.
The second is zero-shot classifier, DetectGPT~\cite{mitchell2023detectgpt}, works under the assumption that AI-generated texts variations typically have lower model probability than the original, while human-written could go either way.
As current zero-shot classifiers require input documents, with considerable length (exceeding 100 tokens) for the classifier to effectively capture contextual features of the text, for which in terms of classifying short sentences, their performances are relatively poor.
Kirchenbauer et al.~\cite{kirchenbauer2023watermark} demonstrated how to incorporate a watermark using only the logarithmic credentials of each step to mark AI-generated texts.
While watermark-based detectors are an intriguing area of research, adding watermarks may affect the readability of the texts, and the removal of watermarks is also a challenge we need to address.
Another noteworthy issue that previous work has focused on is to distinguish whether an entire document is generated by AI.
However, users often use language models to modify portions of texts rather than fully trusting the model to generate an entire document. Therefore, it is also important to explore fine-grained (e.g. sentence-level) detection of AI-generated texts.

Before the era of LLMs, models needed to learn task-specific knowledge and the alignment between task inputs and desired outputs. This is why training on negative samples could sometimes be beneficial, as it provided the model with supplementary knowledge and boundaries for the task-specific information~\cite{li2023quantity}. In the era of LLMs, models no longer need to learn task-specific knowledge and alignment between task inputs and desired outputs, as most of the required knowledge has already been learned during pre-training. Instruction tuning can facilitate the alignment between the model and the expected user task responses. We introduce LLM-Detector, a powerful method to address the challenges of text detection. Specifically, in document-level AI-generated text detection, we label the dataset and use it for Instruction Tuning of LLMs. In sentence-level AI-generated text detection, we label each sentence in the dataset and use it for Instruction Tuning. We also investigate the impact of instruction tuning on text detection performance using text generated by a specific LLM and the influence of different Chinese and English language models on detection performance. Experimental results show that existing methods like Fast-DetectGPT~\cite{bao2023fast}, MPU~\cite{textsmultiscale}, GLTR~\cite{gehrmann2019gltr} are not effective in sentence-level AI-generated text detection. Our proposed LLM-Detector achieves promising results in both sentence and document-level AI-generated text detection challenges and exhibits excellent generalization on OOD datasets. Our contributions are summarized as follows:

\begin{itemize}
    \item To promote research in the field of AI-generated Chinese text detection based on Instruction Tuning for LLMs, particularly to delve into the discrepancies between humans and LLMs, we have compiled 151.7k responses to the same directive questions from human experts. These directive questions span across various domains, including open domains, computer science, finance, medicine, law, psychology, journalism, etc. This dataset contains document-level and sentence-level text annotations and can be used to analyze the characteristic differences in language and style between humans and LLMs, holding significant value for guiding the future development of LLMs in Chinese text detection.
    \item We proposed LLM-Detector, a text detection model that can determine whether text is generated by humans or AI. The model significantly improves the limitations of previous technology in text detection. Specifically, in in-domain detection, LLM-Detector has an accuracy rate of up to 98.52\%, far surpassing statistical-based detectors (such as GLTR's 77.06\% and Fast-DetectGPT's 59.55\%) and supervised classifiers (such as RoBERTa's 89.93\%). In OOD detection, its accuracy rate is 96.70\%, while other detectors' performance significantly decreases. Our experiments indicate that using training data consistent with the target language is crucial for improving the model's performance on specific language text detection tasks. Furthermore, since LLM-Detector is trained based on open-source LLMs, it is easy to customize for deployment.
    \item We conducted sentence-level text detection experiments using a dataset that mixed sentences generated by human experts and AI. The experimental results indicate that existing methods (such as Sent-RoBERTa and Sniffer) face difficulties in solving the problem of sentence-level AI-generated text detection. Our proposed LLM-Detector, however, achieved encouraging results in document-level and cross-domain text detection challenges and demonstrated outstanding generalization capabilities in sentence-level text detection.
\end{itemize}

%%%%%%%%%%%%%%%%%%%%%%%%%%%%%%%%%%%%%%%%%%%%%%%%%%%%%%%%%%%%%%%%%%%%%%%%%%%%%%%
%%%%%%%%%%%%%%%%%%%%%%%%%%%%%%%%%%%%%%%%%%%%%%%%%%%%%%%%%%%%%%%%%%%%%%%%%%%%%%%
% Related Work
% Check by Rongsheng Wang,
%%%%%%%%%%%%%%%%%%%%%%%%%%%%%%%%%%%%%%%%%%%%%%%%%%%%%%%%%%%%%%%%%%%%%%%%%%%%%%%
%%%%%%%%%%%%%%%%%%%%%%%%%%%%%%%%%%%%%%%%%%%%%%%%%%%%%%%%%%%%%%%%%%%%%%%%%%%%%%%
\section{Related Work}
\label{Related Work}
LLMs have been pre-trained on extensive text corpora, enabling them to generate contextually relevant and fluent texts. However, this also increases the difficulty of detecting AI-generated texts. The existing methods for detecting generated texts can be broadly categorized into two types: black-box and white-box detection~\cite{tang2023science}, contingent upon the level of access to the model that is suspected to have generated the target texts.

\subsection{Black-Box Detection}
\label{Black-Box Detection}
For black-box detection, classifiers are restricted to API-level access to LLMs (only available for the text).
To develop a proficient detector, black-box methods are typically designed to first extract and select features based on text samples. 
Originating from both human and AI-generated texts, the black-box detection method would train a classification model leveraging relevant features, for which heavily relies on the large amount of text data and detectors.

\textbf{Datasets.} Recently, a growing body of research has concentrated on amassing responses generated by LLMs and comparing them to human-written texts spanning a wide range of domains.
~\cite{guo2023close} collected the HC3 (Human ChatGPT Comparison Corpus) Chinese dataset, which consists of nearly 40K questions and their corresponding answers from human experts and ChatGPT, which has a wide range of domains coverage (open-domain, computer science, finance, medicine, law, and psychology).
~\cite{wang2023m4} collected the M4 (Multi-generator, Multi-domain, and Multi-lingual Black-Box Machine-Generated Text Detection) dataset, which consists of questions and their corresponding answers from human experts and LLMs, covering a wide range of languages (English, Chinese, Russian, Arabic, Indonesian and Urdu).
Overall, Previous work has not established a comprehensive Chinese text detection dataset that encompasses diverse Chinese text data with different parameter types from LLMs and human expert responses.

\textbf{Detectors.} Existing black-box detectors can be grouped into two main categories: supervised classifiers and zero-shot classifiers. 
% Supervised classifiers are more commonly used in AI-generated text detection.
Logistic regression with GLTR~\cite{gehrmann2019gltr} features and an end-to-end RoBerta~\cite{guo2023close} classifier, to detect whether a certain text (English and Chinese) is generated by ChatGPT or humans across several domains.
However, the study conducted demonstrates that a limitation of supervised models is the potential occurrence of overfitting within the domain, resulting in poor detection performance OOD (OOD)~\cite{chakraborty2023possibilities}.
To address the limitations of supervised classifiers, zero-shot classifiers, using a pre-trained language model directly without fine-tuning, are immune to domain-specific degradation.
Zero-shot classifiers such as GPT-Zero~\footnote{\url{https://gptzero.me/}}, DetectGPT~\cite{mitchell2023detectgpt} and Fast-DetectGPT~\cite{bao2023fast} have been developed.
These methods utilize checks on perplexity and burstiness in the text to determine whether it is artificially generated or authored by a human.
The current zero-shot classifiers require input documents of considerable length (exceeding 100 tokens) for the classifier to effectively capture contextual features of the text.
In terms of classifying short sentences, their performance is relatively poor.

\subsection{White-Box Detection}
\label{White-Box Detection}
White-box detection require fully access to LLMs, thereby enabling control over the generation behavior of the model or embedding watermark within the generated texts. This enables the tracking and detection of AI-generated texts within white-box settings. 

White-Box detection involves using statistical boundaries between linguistic patterns found in human-written and AI-generated text as proxies.
These boundaries are determined based on $n$-gram frequencies~\cite{badaskar2008identifying}, entropy~\cite{lavergne2008detecting}, and perplexity~\cite{beresneva2016computer}.
One limitation of these statistics-based methods is the assumption, which assumes access to the model's prediction distributions.
This constraint hinders broader applications, especially for models behind APIs.

Inspired by copyright protection watermarks in the image and video fields, as proposed by~\cite{kirchenbauer2023watermark}, partitions the model’s vocabulary into whitelist and black list tokens when predicting the next token given a prompt.
During text generation, the goal is to produce whitelist tokens as much as possible, effectively creating a strong watermark.
The third-parties can determine if the text is machine-generated by analyzing the frequency of whitelist tokens within the text. While watermarking methods offer robustness and interpretability, they can compromise the quality of the generated text and may not be highly practical in certain scenarios.

%%%%%%%%%%%%%%%%%%%%%%%%%%%%%%%%%%%%%%%%%%%%%%%%%%%%%%%%%%%%%%%%%%%%%%%%%%%%%%%
%%%%%%%%%%%%%%%%%%%%%%%%%%%%%%%%%%%%%%%%%%%%%%%%%%%%%%%%%%%%%%%%%%%%%%%%%%%%%%%
% Methodology
% Check by Rongsheng Wang,
%%%%%%%%%%%%%%%%%%%%%%%%%%%%%%%%%%%%%%%%%%%%%%%%%%%%%%%%%%%%%%%%%%%%%%%%%%%%%%%
%%%%%%%%%%%%%%%%%%%%%%%%%%%%%%%%%%%%%%%%%%%%%%%%%%%%%%%%%%%%%%%%%%%%%%%%%%%%%%%
\section{Methodology}
\label{Methodology}

\subsection{Overview of LLM-Detector}
\label{Overview of LLM-Detector}
\begin{figure*}[ht]
\vskip 0.2in
    \centering
    \includegraphics[width=\linewidth]{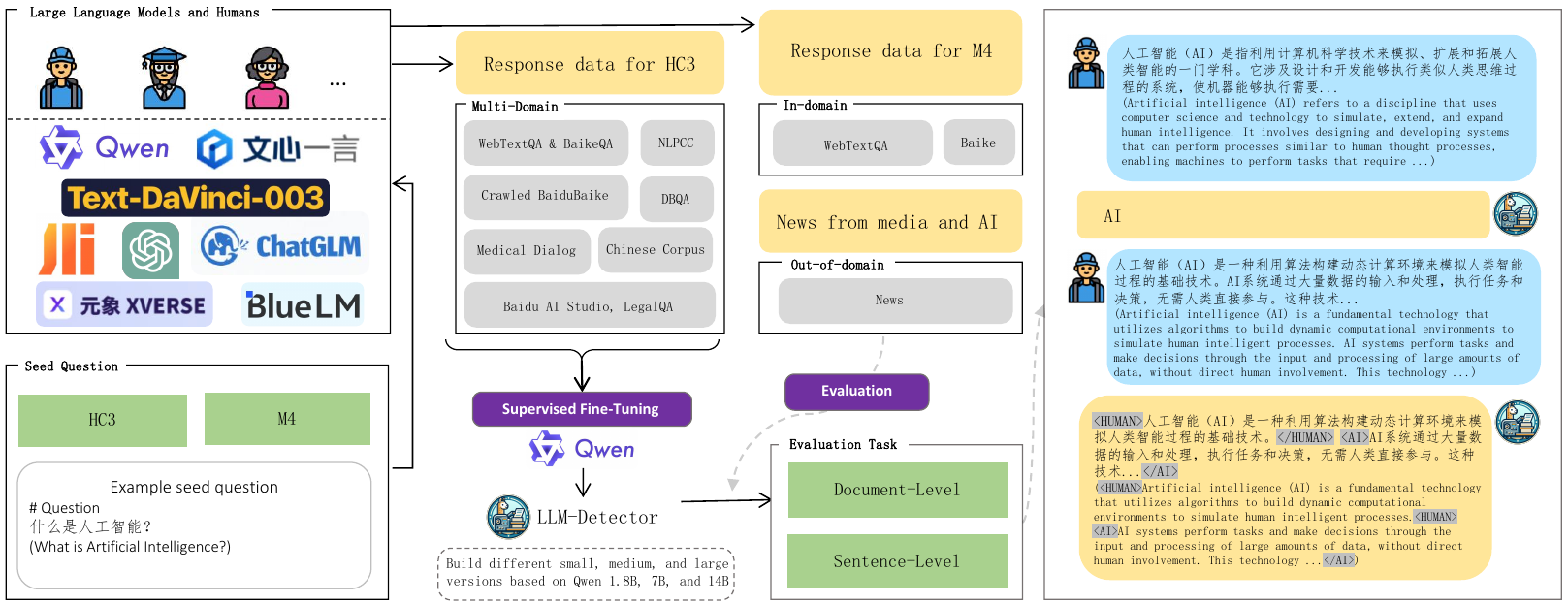}
    \caption{LLM-Detector Framework. First, HC3 and M4 seed questions are used to prompt responses from human experts and multiple LLMs, where the responses to HC3's multi-domain seed questions will be employed to train the LLM-Detector. Second, the responses generated using M4 seed questions from the same domain as HC3 are utilized to test the in-domain capabilities of the LLM-Detector, while an additionally constructed News dataset is used to test the LLM-Detector's OOD capabilities.}
    \label{fig:model}
    \vskip -0.2in
\end{figure*}

The structure of our proposed Chinese text detection model, LLM-Detector, is shown in Figure~\ref{fig:model}. In the training stage, we constructed a response dataset based on HC3 seed questions, which consists of responses generated by human experts and multiple LLMs, including their source labels (AI or human) and more granular sentence-level annotations. These sentence-level annotations include mixed texts written by humans and polished by AI. Subsequently, we adapted a foundational LLM to LLM-Detector through instruction tuning, fine-tuning it on response samples from human experts and multiple LLMs to elicit the model's Chinese text detection capabilities. 
In the evaluation stage, we input the corresponding instruction text into the LLM-Detector for detection based on the joint responses generated by M4 seed problems from LLMs and human experts. The diverse of the LLM-Detector's Chinese text detection dataset provides better guidance for the foundational LLM in modeling the connection between user instructions and appropriate responses, thereby enhancing the text detection capabilities of the instruction-tuned LLM.

The organization of the second part of this document is as follows: Section~\ref{Generating detection data with different LLMs} provides a detailed description of the process of building Chinese text detection data by humans and multiple LLMs; Section~\ref{Design of LLM-Detector} explains the design of LLM-Detector; and finally, Section~\ref{Human experts and LLMs constructing test set} introduces the in-domain and OOD datasets we created. Section~\ref{Dataset Overview} provides an overview of all datasets used in this work and compares the differences between human-written and AI-generated text.

\subsection{Generating detection data with different LLMs}
\label{Generating detection data with different LLMs}

\subsubsection{Generation of Document-Level Data}
HC3 \cite{guo-etal-2023-hc3} is the first human-ChatGPT comparison corpus which contains 12, 853 questions from WebText Q\&A, Baike Q\&A, Medical Dialog, Chinese Corpus, Legal Q\&A, etc.

Specifically, We further utilized the 12,853 sub-questions from HC3, allowing 9 different LLMs (including ChatGPT, GPT-4, etc.) to generate responses, which we labeled as "AI". All LLMs used are displayed in Figure~\ref{fig:model}. Finally, we combined the original human expert responses from HC3 with the newly generated responses to create a training dataset. According to the experimental design, we filter texts with a length of less than 10 as the final training set. An example of the generated document-level detection data can be found in Appendix ~\ref{app:Example of document-level data organization}. 
% The details of the training set can be shown in Appendix~\ref{app:Dataset Source Parse Analysis} and Table~\ref{tab:doc_level_data_size}.

\subsubsection{Generation of Sentence-Level Data}
To construct the sentence-level dataset, we sampled 5,589 human responses from the dataset provided by HC3 as the data source. Then we collect longer sentence lengths from the data source and use regular expressions to break sentences. In addition, we randomly selected several sentences whose number could be in [1, len\_of\_the\_sents-1]  to ensure that the text contains at least one human sentence and input them into a large language model for polishing. The specific process of generating sentence-level data is detailed in the Appendix\ref{app:Example of sentence-level data organization}.

% \begin{table}[ht]
%     \caption{Size of sentence-level dataset.}
%     \label{tab:sent_level_data_size}
%     \centering
%     \begin{tabular}{ccc}
%         \hline
%          & Total \\ \hline
%         Sentence-level Train Set & 5,589\\ \hline
%         Sentence-level Test Set & 1,504\\ \hline
%     \end{tabular}
% \end{table}

\subsection{Design of LLM-Detector}
\label{Design of LLM-Detector}

Specifically, given an instruction dataset $V$ of instruction pairs $x$ = (\textsc{Instruction}, \textsc{Output}) with $x\in V$, each instruction $x$ is generated by either human experts or LLMs, and is labeled as $x_r$ according to its source. The text detection dataset $R$ is ultimately formed, which includes instruction pairs with their corresponding source labels (Human or AI), represented as $R=\{(x,x_r)\ |\ x\in V\}$. During the coach instruction tuning process, each $(x,x_r)\in R$ is leveraged to construct an instruction pair $x_c$, leading to an instruction dataset $C=\{x_c\ |\ x\in V\}$.

Table~\ref{tab:instruction} illustrates how the \textsc{Instruction} of $x_c$ guides LLM to detect the text source of $x$ (the original instruction pair), with the \textsc{Output} of $x_c$ being $x_r$, which is the label for the text source. When constructing the \textsc{Instruction} component, we designed a concise and clear detection instruction that explicitly indicates LLM should learn the text detection task based on the text source label and the instruction text. To prevent the distraction that might arise from lengthy instructions, we deliberately did not create an exhaustive instruction to cover all criteria, allowing LLM to better focus on the relationship between the input instruction text pairs and their corresponding text source labels.

\begin{table}[ht]
\centering
\vspace{-0.5cm}
\caption{Illustration of the format for the instruction pairs $x_c$ in coach instruction tuning. $x$ represents the instruction text, while $x_r$ denotes the label for the text source.}
\label{tab:instruction}
\begin{tcolorbox}
\textbf{Instruction}: Categorize the texts into one of the 2 classes: human or AI. \textbf{Input}[$x$] \\
\textbf{Output}: [$x_r$]
\end{tcolorbox}
\end{table}

Given an LLM with parameters $\theta$ as the initial model for coach instruction tuning, training the model on the constructed instruction dataset $C$ results in the adaptation of the LLM's parameters from $\theta$ to $\theta_c$, denoted as LLM-Detector. Specifically, $\theta_c$ is obtained by maximizing the probability of predicting the next tokens in the \textsc{Output} component of $x_c$, conditioned on the \textsc{Instruction} of $x_c \in C$, which is formulated as follows:
\vspace{-0.2cm}
\begin{equation}
   \theta_c = \argmax_{\theta} \sum\limits_{x_c\in C}\log P(\textsc{Output}\,|\,\textsc{Instruction};\theta,x_c). \label{eq1}
\end{equation}

\subsection{Human experts and LLMs constructing test set}
\label{Human experts and LLMs constructing test set}
To evaluate the detection ability of our method, we construct an in-domain test set and an OOD test set.

\subsubsection{Generation of Document-Level In-Domain Data}
M4 \cite{wang2023m4} is a large-scale benchmark that is multigenerator, multi-domain, and multi-lingual corpus for machine-generated text detection. 
The domains of M4 include Wikipedia, WikiHow, Reddit, arXiv, RuATD, Baike. We sampled Chinese language questions from M4 and generated responses using nine different LLMs (including ChatGPT, GPT-4, etc.). All LLMs used are displayed in Figure~\ref{fig:model}. Finally, we combined the original human expert responses from M4 with the newly generated responses to create a test dataset. Similar to building a training set, we also filter sentences with a length of less than 10 as the final in-domain test set.

Because the sources of HC3 and M4 contain the same public corpus as their sources like the large scale Chinese corpus for NLP \cite{bright_xu_2019_3402023}, we use M4 as the in-domain test set to evaluate the performance of our method.
% The details of the in-domain test set can be shown in Appendix \ref{app:Dataset Source Parse Analysis} and Table \ref{tab:doc_level_data_size}.

\subsubsection{Generation of Document-Level Out-of-Domain Data}

News Broadcast Text~\footnote{\url{https://cn.govopendata.com/xinwenlianbo/}} is the text version of News Broadcast which is crawled through the public data of the CCTV network. Besides, we generated news in some fields through ChatGPT, such as sports and science. 

In this process, we use the temperature 0.7, set the top\_p to 1, and adopt the max\_tokens as 4096, others are the default. 
The prompt that we give to ChatGPT can be found in the appendix~\ref{app:The prompt for constructing the OOD dataset}.
% The details of the OOD test set can be shown in Appendix \ref{app:Dataset Source Parse Analysis} and Table \ref{tab:doc_level_data_size}. 

% \subsubsection{Generation of M4 Sentence-Level Out-of-Domain Text Detection Dataset}

% To evaluate our method, we construct the sentence-level text detection dataset by Human and AI.
% Some examples can be shown as follows: 
% \begin{lstlisting}
% [{
%     "instruction": "S1 S2 S3 S4 S5",
%     "input": "",
%     "output": "<HUMAN>S1</HUMAN><HUMAN>S2</HUMAN><AI>S3</AI><AI>S4</AI><AI>S5</AI>"
% }]
% \end{lstlisting}

\subsection{Dataset Overview}
\label{Dataset Overview}
In order to address the challenges mentioned above, we first constructed a multi-domain dataset for Chinese text detection, which includes responses from different LLMs and human experts.
These responses were generated by \textbf{9 types of LLMs}, including ChatGPT, GPT-4, etc.
The data sources cover \textbf{7 main domains}, including web Q\&A, encyclopedias, Baidu-Baike, medical dialogues, etc., for which with \textbf{151.7k} data samples, each recorded according to its generation source.
For the 151.7k data mentioned above, we used 118.1k data samples for Instruction Tuning LLMs, and 29.7k data samples were used to evaluate the detection performance in the in-domain.
The remaining data samples were from news data of different domains than the previous training and evaluation datasets, used to evaluate the model's OOD detection performance.
Again, we built a sentence-level AI-generated Chinese text detection dataset, which includes both human-written sentences and AI-generated sentences, more likely to appear in real AI-assisted writing.
This dataset contains a total of \textbf{7.1k} data samples, as shown in Appendix \ref{app:Details of the document-level dataset}, Appendix~\ref{app:Dataset Source Parse Analysis} and Table~\ref{tab:doc_level_data_size}. We do a linguistic and semantic analysis for our dataset as shown in Appendix \ref{app:Detailed analysis of the dataset}.

% Our dataset encompasses a variety of fields and responses from multiple LLMs, with the specific sources of the responses detailed as shown in Appendix \ref{app:Details of the document-level dataset}, Appendix~\ref{app:Dataset Source Parse Analysis} and Table~\ref{tab:doc_level_data_size}. We do a linguistic and semantic analysis for our dataset as shown in Appendix \ref{app:Detailed analysis of the dataset}.

% Our dataset can be shown as follows:
% \begin{lstlisting}
% [{
%     "instruction": "S1 S2 S3 S4 S5",
%     "input": "",
%     "output": "L1"
% }]
% \end{lstlisting}

% Our dataset was constructed from 20 files generated by human and different large language models such as GPT-4, ChatGLM2 \cite{zeng2022glm}, Qwen \cite{qwen}, Baichuan2 \cite{baichuan2023baichuan2}, as shown in Figure \ref{fig:2}.

\begin{table}[ht]
    \centering
    \caption{Size of document-level Training Set and Test Set.}
    \label{tab:doc_level_data_size}
    \centering
    \begin{tabular}{ccc}
        \hline
         & Data Source & Total \\ \hline
        \multirow{2}{*}{Train Set}
        & Human & 21,681 \\ 
        & AI & 96,453 \\ \hline
        \multirow{2}{*}{In-Domain Test Set}
        & Human & 3,000 \\ 
        & AI & 26,750 \\ \hline
        \multirow{2}{*}{Out-of-Domain Test Set} & Human & 2,000 \\ 
        & AI & 1,915 \\ \hline
        Total & - & 151,799 \\ \hline
    \end{tabular}
\end{table}

In the training set, human sentences account for around 18.4\%, while AI sentences account for about 81.6\%. The dataset is mainly based on sentences generated by AI because we mainly want the model to learn the features of AI sentences. However, the model also needs human sentences as a comparison to learn the different features of the two generation mechanisms for improving the detection ability of the model.

%%%%%%%%%%%%%%%%%%%%%%%%%%%%%%%%%%%%%%%%%%%%%%%%%%%%%%%%%%%%%%%%%%%%%%%%%%%%%%%
%%%%%%%%%%%%%%%%%%%%%%%%%%%%%%%%%%%%%%%%%%%%%%%%%%%%%%%%%%%%%%%%%%%%%%%%%%%%%%%
% Experiments
%%%%%%%%%%%%%%%%%%%%%%%%%%%%%%%%%%%%%%%%%%%%%%%%%%%%%%%%%%%%%%%%%%%%%%%%%%%%%%%
%%%%%%%%%%%%%%%%%%%%%%%%%%%%%%%%%%%%%%%%%%%%%%%%%%%%%%%%%%%%%%%%%%%%%%%%%%%%%%%
\section{Experiments}

\subsection{Tasks}
Previous work has mostly focused on document-level AI-generated text detection~\cite{mitchell2023detectgpt, guo2023close}, and many of these efforts are difficult to extend to sentence-level AI-generated text detection. Additionally, previous methods often exhibit limitations on OOD~\cite{chakraborty2023possibilities} datasets. To address this, we have defined a variety of tasks to test the performance of the LLM-Detector. This includes in-domain text detection, OOD text detection, and sentence-level text detection. The specific descriptions of the three tasks can be found in appendix~\ref{app:Details of Tasks}.

\subsection{Baselines}
For zero-shot classifiers, we mainly compare our proposed LLM-Detector with Fast-DetectGPT~\cite{bao2023fast}, GLTR~\cite{gehrmann2019gltr}, PPL~\cite{guo2023close}, ChatGPT and GPT-4. For the supervised classifier, we also conducted fine-tuning training on Bert~\cite{devlin2018bert}, RoBerta~\cite{liu2019roberta}, MPU~\cite{textsmultiscale}, LLaMA2~\cite{touvron2023llama}, and Mixtral~\cite{jiang2023mistral} using our own dataset. And we compared it with LLM-Detector. A detailed description of these previous methods can be found in the appendix~\ref{app:Details of Baselines}.

\subsection{Experimental Settings}

\textbf{Computational resources and parameter settings.} Our model is built upon Qwen~\cite{bai2023qwen}, a Chinese large language model with parameter sizes of 1.8 billion, 7 billion, and 13 billion. Based on Qwen LLM with different parameters, we trained Small, Medium, and Large LLM-Detectors to distinguish the impact of model parameters on accuracy. The training process utilizes 4 A100 (80G) GPUs with parallelization, incorporating quantized low-rank adaptation (QLoRA)~\cite{dettmers2023qlora}. This methodology is implemented through transformers and PEFT libraries. To manage training costs, we employ fp16 precision with ZeRO-2~\cite{rajbhandari2021zero}, a gradient accumulation strategy. During the entire training process, the learning rate is 5e-5, the training epochs are 3.0, and the LoRA Rank is 8. At the end of the entire training process, the best model was saved for evaluation. For the training of BERT, RoBERTa, and MPU, the learning rate is 1e-3 and the training epochs are 50.0.

\textbf{Metrics.} We utilize accuracy (ACC) to assess the performance of models in classifying text, distinguishing between human-written and AI-generated content. For sentence-level evaluation, we utilize metrics include precision (P.), recall (R.), and Macro-F1. Precision and recall individually represent the "accuracy" and "coverage" of each category. The Macro-F1 Score serves as an effective combination of these two indicators, providing a comprehensive measure of overall performance. A detailed description of these previous methods can be found in the Appendix~\ref{app:Details of Metrics}.

\subsection{Main Results}
The experimental results are shown in Table~\ref{tab:id_acc}. Generally, after fine-tuning with LLM-Detector, models of various sizes can achieve significantly better performance. This demonstrates the effectiveness and broad applicability of instruction tuning for large models in enhancing LLM text detection. In addition, we have the following observations.

% \vspace{-1.5em}
% \vspace{0.1 mm}
% \vspace{-1.6cm}

\begin{table}[ht]
  \centering
  \vspace{-0.5cm}
  \caption{Experimental results of different detection models on the in-domain dataset. "finetuned" indicates models that have been trained on the same dataset. Bold text indicates the model with the best performance on the in-domain dataset.}
  \begin{tabular}{lc}
    \toprule
    % \cmidrule(r){1-2}
    Model & Accuracy \\
    \midrule
    \multicolumn{2}{c}{\cellcolor[gray]{0.8}\textbf{Statistical-based Classifier}} \\
    Fast-DetectGPT    & 59.55\% \\
    GLTR & 77.06\% \\
    PPL    & 10.26\% \\
    \multicolumn{2}{c}{\cellcolor[gray]{0.8}\textbf{Zero-Shot Classifier}} \\
    ChatGPT & 81.46\% \\
    GPT-4   & 37.41\% \\
    \multicolumn{2}{c}{\cellcolor[gray]{0.8}\textbf{Supervised Classifier}} \\
    BERT-finetuned   & 76.50\% \\
    RoBERTa-finetuned     & 89.93\% \\
    BERT-MPU    & 75.95\% \\
    RoBERTa-MPU     & 89.93\% \\
    LLaMA-2-7B-finetuned     & 83.65\% \\
    LLaMA-2-13B-finetuned      & 96.53\% \\
    Mistral-7B-finetuned      & 97.98\% \\
    \midrule
    LLM-Detector-Small   & 97.84\% \\
    LLM-Detector-Medium   & 98.35\% \\
    \rowcolor[RGB]{135,206,250} \textbf{LLM-Detector-Large}   & \textbf{98.52\%} \\
    \bottomrule
  \end{tabular}
  \label{tab:id_acc}
\end{table}

On the in-domain dataset, classifiers based on supervised learning typically outperform zero-shot and statistical-based classifiers. Furthermore, classifiers trained on large language models surpass those based on smaller parameter models such as BERT and RoBERTa. This confirms the inherent advantage of large-scale parameters in model performance. On the other hand, there is a positive correlation between the scale of the model and its detection performance. For instance, the LLM-Detector-Large model outperforms the LLM-Detector-Medium and LLM-Detector-Small models in terms of accuracy. The LLaMA-13B model has higher accuracy than the LLaMA-7B model.

When fine-tuning large language models for text detection tasks, Chinese LLMs trained on Chinese data significantly outperform those trained on English data. This phenomenon indicates that using training data consistent with the target language is crucial for improving the performance of models on specific language tasks. For instance, when fine-tuned with Chinese LLMs (such as Qwen), the resulting LLM-Detector typically achieves higher accuracy than its counterparts based on English LLMs (such as Mistral and LLaMA).

We investigate the performance of various detection models on out-of-distribution (OOD) datasets, The results are shown in Table~\ref{tab:ood_acc}. The experimental results indicate that among the statistical classifier models, the Fast-DetectGPT model achieves the highest accuracy, reaching 94.48\%. In the supervised classifier models, the LLaMA-2-13B-finetuned model has the highest accuracy, achieving 93.19\%. Among all models, the LLM-Detector-Large model demonstrates the most impressive performance, with an accuracy of 96.70\%. These results suggest that for the detection of OOD datasets, the LLM-Detector-Large model is a choice with high accuracy and effectiveness.

\begin{table}[ht]
 \caption{Experimental results of different detection models on the OOD dataset. "finetuned" indicates models that have been trained on the same dataset. Bold text and blue background indicates the model with the best performance on the OOD dataset.}
  \centering
  \begin{tabular}{lc}
    \toprule
    % \cmidrule(r){1-2}
    Model & Accuracy \\
    \midrule
    \multicolumn{2}{c}{\cellcolor[gray]{0.8}\textbf{Statistical-based Classifier}} \\
    Fast-DetectGPT    & 94.48\% \\
    GLTR & 78.60\% \\
    PPL    & 51.09\% \\
    \multicolumn{2}{c}{\cellcolor[gray]{0.8}\textbf{Supervised Classifier}} \\
    BERT-finetuned   & 48.39\% \\
    RoBERTa-finetuned     & 48.95\% \\
    BERT-MPU    & 23.07\% \\
    RoBERTa-MPU     & 48.95\% \\
    LLaMA-2-7B-finetuned     & 87.05\% \\
    LLaMA-2-13B-finetuned      & 93.19\% \\
    Mistral-7B-finetuned      & 92.73\% \\
    \midrule
    LLM-Detector-Small   & 90.22\% \\
    LLM-Detector-Medium   & 93.57\% \\
    \rowcolor[RGB]{135,206,250} \textbf{LLM-Detector-Large}   & \textbf{96.70\%} \\
    \bottomrule
  \end{tabular}
  \label{tab:ood_acc}
\end{table}

We implement Sniffer and Sent-RoBERTa based on sentence-level detection tasks, as well as our LLM-Detector. Sniffer~\cite{li2023origin} is a powerful model that can detect and trace the origins of AI-generated texts. To perform sentence-level detection, we train a sentence-level Sniffer following the structure and training process of the original Sniffer, but using a single sentence as input instead of an entire document. RoBERTa~\cite{liu2019roberta} is built on the Transformer encoder and can handle both sentence classification tasks and sequence labeling tasks. We train a sentence-level RoBERTa for detection using a method based on sentence classification. The results in Table~\ref{tab:diff_data_prf} clearly show that our LLM-Detector outperforms the other two methods, demonstrating its effectiveness. In contrast, Sent-RoBERTa's performance is noticeably inferior, highlighting the challenge of adapting document-level detection methods for sentence-level detection.

\begin{table}[ht]
 \caption{Performance comparison of different LLMs based on different dataset sources.}
  \centering
  \begin{tabular}{cccc}
    \toprule
     Model & P. & R. & Macro-F1 \\
    \midrule
    Sniffer & 65.0\% & 64.24\% & 62.51\% \\
    Sent-RoBERTa & 37.30\% & 42.15\% & 39.11\% \\
    \midrule
    \rowcolor[RGB]{135,206,250} \textbf{LLM-Detector-Small} & \textbf{71.36\%} & \textbf{72.62\%} & \textbf{73.5\%}\\
    \bottomrule
  \end{tabular}
  \label{tab:diff_data_prf}
\end{table}

\subsection{Usability Analysis}

\textbf{Robustness on Short Texts.} Zero-shot detectors, due to their statistical properties, are expected to perform worse on shorter texts, and similarly, supervised learning detectors will also face the same issue. We sampled texts with lengths ranging from 10 to 50 from the in-domain dataset to evaluate different detectors. As shown in Table~\ref{tab:shot_text_acc}, the LLM-Detector and text detectors trained based on LLMs (such as Mistral and LLaMA) do not exhibit a significant decrease in accuracy with shorter text lengths; that is, the detection accuracy does not typically decrease due to the brevity of the text. In contrast, supervised detectors see a substantial decrease in detection accuracy on short texts. We speculate that this is because supervised detectors are unable to effectively capture the characteristics of human- and AI-generated texts in contexts where there is insufficient context. Statistical-based detectors also experience a decrease in accuracy when detecting short texts.

\begin{table}[ht]
 \caption{Accuracy of different detectors on Short Texts."finetuned" indicates models that have been trained on the same dataset. Bold text and blue background indicates the model with the best performance on the short texts.}
  \centering
  \begin{tabular}{lc}
    \toprule
    % \cmidrule(r){1-2}
    Model & Accuracy \\
    \midrule
    \multicolumn{2}{c}{\cellcolor[gray]{0.8}\textbf{Statistical-based Classifier}} \\
    Fast-DetectGPT    & 72.78\% \\
    GLTR & 65.36\% \\
    PPL    & 57.03\% \\
    \multicolumn{2}{c}{\cellcolor[gray]{0.8}\textbf{Zero-Shot Classifier}} \\
    ChatGPT  & 67.26\% \\
    GPT-4    & 57.16\% \\
    \multicolumn{2}{c}{\cellcolor[gray]{0.8}\textbf{Supervised Classifier}} \\
    BERT-finetuned    & 42.02\% \\
    RoBERTa-finetuned     & 43.25\% \\
    BERT-MPU    & 37.88\% \\
    RoBERTa-MPU     & 43.25\% \\
    LLaMA-2-7B-finetuned     & 96.38\% \\
    LLaMA-2-13B-finetuned      & 97.94\% \\
    Mistral-7B-finetuned      & 98.87\% \\
    \midrule
    LLM-Detector-Small   & 97.80\% \\
    LLM-Detector-Medium   & 98.80\% \\
    \rowcolor[RGB]{135,206,250} \textbf{LLM-Detector-Large}   & \textbf{99.20\%} \\
    \bottomrule
  \end{tabular}
  \label{tab:shot_text_acc}
\end{table}

We further investigated the impact of text length on Fast-DetectGPT and RoBERTa, as shown in the Appendix~\ref{app:The impact of text length on Fast-DetectGPT and RoBERTa}. We continued to sample texts of lengths 100, 150, and 200 from the in-domain dataset for detection. As the text length increased gradually, both Fast-DetectGPT (a statistical detector) and RoBERTa (a supervised detector) saw improvements in accuracy. After the text length exceeded 100 characters, the accuracy of Fast-DetectGPT rapidly rose to 94.3\%. When the text length exceeded 200 characters, the accuracy of RoBERTa rapidly increased to 83.8\%.

% \begin{figure}[H]
% \vskip 0.1in
%   \centering
%   \includegraphics[width=0.95\linewidth]{icml2024/accuracy_performance.pdf}
%   \caption{As the length of the text increases, the accuracy performance of Fast-DetectGPT and RoBERTa.}
%   \label{fig:text_len}
%   \vskip -0.1in
% \end{figure}

\begin{figure}[ht]
\vskip 0.1in
  \centering
  \includegraphics[width=\linewidth]{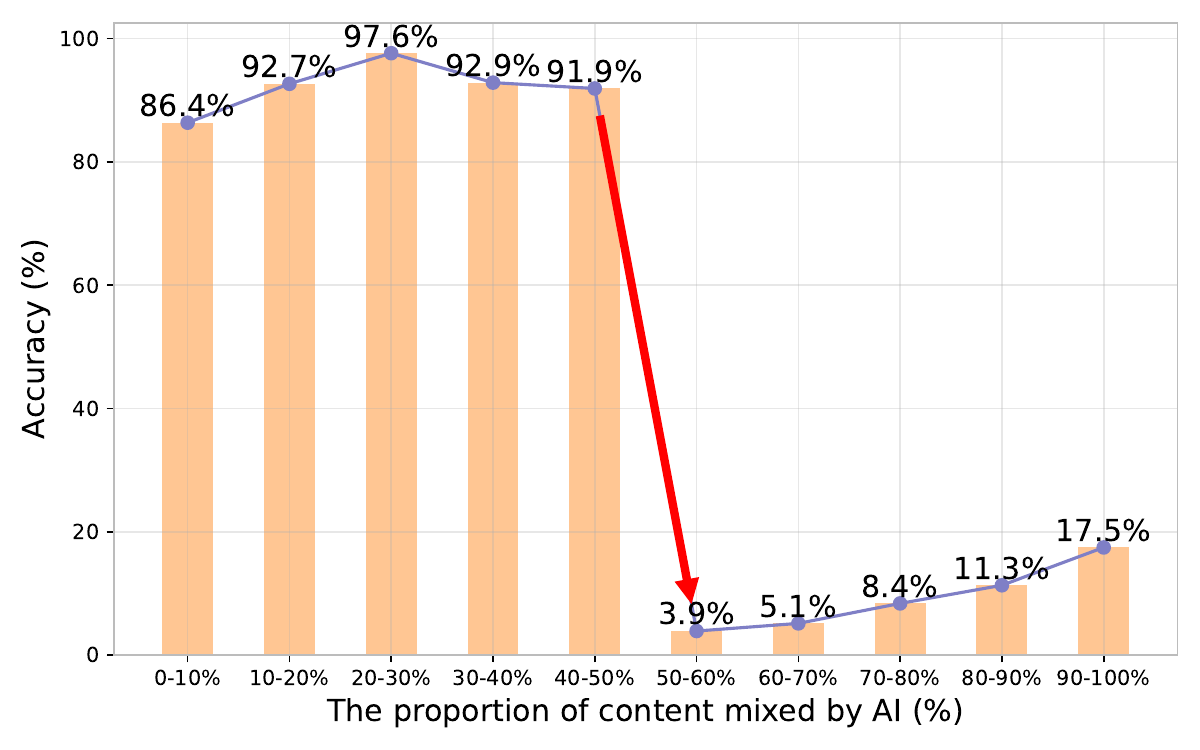}
  \caption{The accuracy performance increases with the proportion of AI-generated content.}
  \label{fig:ai_cont_prop}
  \vskip -0.1in
\end{figure}

\textbf{Robustness in Mixed Text Detection.} In further research, we explored the impact of mixed text on the performance of LLM-Detector, and the results are shown in Figure~\ref{fig:ai_cont_prop}. We found that when the proportion of mixed text reaches 50\%-60\%, the detection accuracy drops sharply. This is because, as the proportion of mixed text increases, the characteristics of the original text may be weakened. For LLM-Detector, this may mean that the signals used to judge the authenticity of the text become weaker, while noise increases, leading to a decline in model performance. Overall, LLM-Detector exhibits a certain robustness in mixed text detection and is able to resist the influence of mixed text to a certain extent.

\textbf{Are instruction-tuned LLMs better at detecting text they themselves have generated?} To evaluate the detection performance of different LLMs on content they have generated themselves, we fine-tuned the LLMs on responses generated by three different LLMs and human-written texts. The results are shown in the appendix~\ref{app:Are instruction-tuned LLMs better at detecting text they themselves have generated}. A notable trend is that LLMs tend to perform best on text detection with texts they have generated. For instance, the ChatGLM2-6B model achieves the highest accuracy (99.91\%) on the dataset it generated, which is significantly higher than any other model tested against the same dataset. Similarly, the Qwen-14B model also has a high accuracy of 96.18\% on its generated dataset. However, an interesting anomaly arises with the BlueLM-7B model. The Qwen-7B model outperforms the BlueLM-7B on its own dataset, with an accuracy of 97.8\% compared to 97.1\% for the BlueLM-7B. While this could suggest a potential issue with the BlueLM-7B model's training, it is also worth noting that the difference is very small (only 0.7\%), which could fall within the margin of error.

\textbf{The impact of text generated by LLMs of different scales on the accuracy of text detection.} We used the LLM-Detector to perform text detection on texts generated by LLMs of different parameter sizes. We found that the texts produced by LLMs of varying scales had no significant impact on the accuracy of text detection by LLM-Detector, indicating that detectors trained on LLMs demonstrate better robustness and generalization, as shown in the appendix~\ref{app:The impact of text generated by LLMs of different scales on the accuracy of text detection}. Specifically, the three differently sized detectors—Small, Medium, and Large showed a small range of fluctuation in detection accuracy for texts generated by LLMs of different scales, with the gap between the highest and lowest accuracy not exceeding 5\%.

%%%%%%%%%%%%%%%%%%%%%%%%%%%%%%%%%%%%%%%%%%%%%%%%%%%%%%%%%%%%%%%%%%%%%%%%%%%%%%%
%%%%%%%%%%%%%%%%%%%%%%%%%%%%%%%%%%%%%%%%%%%%%%%%%%%%%%%%%%%%%%%%%%%%%%%%%%%%%%%
% Conclusion
%%%%%%%%%%%%%%%%%%%%%%%%%%%%%%%%%%%%%%%%%%%%%%%%%%%%%%%%%%%%%%%%%%%%%%%%%%%%%%%
%%%%%%%%%%%%%%%%%%%%%%%%%%%%%%%%%%%%%%%%%%%%%%%%%%%%%%%%%%%%%%%%%%%%%%%%%%%%%%%
\section{Conclusion}
In this paper, we have designed a simple yet effective method to detect text generated by AI. Our proposed method is based on the intuition that LLM has learned a wealth of knowledge during the pre-training, which enables LLM to autonomously detect the text it generates. Instruction tuning can facilitate the alignment between the model and the user's expected text detection task responses. Compared to previous methods, our method can accomplish AI-generated text detection at both the document and sentence levels and maintain good performance on OOD data. Therefore, our method possesses superior generalization ability and practicality. We conducted experiments on the three proposed datasets, which cover responses generated by different LLMs, including in-domain and OOD data, and provide more fine-grained sentence-level annotations. The experimental results demonstrate that our method can effectively identify texts generated by LLMs. Moreover, our method shows strong robustness against biases in mixed AI texts, short texts, and OOD texts.

\section*{Impact Statements}
% https://medium.com/@icml2024pc/impact-statements-in-icml-submissions-0ba1ea22119b
% Authors are required to include a statement of the potential broader impact of their work, including its ethical aspects and future societal consequences. This statement should be in a separate section at the end of the paper (co-located with Acknowledgements, before References), and does not count toward the paper page limit. In many cases, where the ethical impacts and expected societal implications are those that are well established when advancing the field of Machine Learning, substantial discussion is not required, and a simple statement such as:
% “This paper presents work whose goal is to advance the field of Machine Learning. There are many potential societal consequences of our work, none which we feel must be specifically highlighted here.”
% The above statement can be used verbatim in such cases, but we encourage authors to think about whether there is content which does warrant further discussion, as this statement will be apparent if the paper is later flagged for ethics review.

% This paper presents work whose goal is to advance the field of Machine Learning. There are many potential societal consequences of our work, none which we feel must be specifically highlighted here.
The proposed AI text detection method offers advancements in content moderation and information security by identifying AI-generated text, ensuring authenticity and reliability. However, it faces challenges such as potential false positives and negatives, and biases inherent in pre-trained LLMs. As AI models evolve, there is a need for ongoing research to enhance the method’s robustness and address these limitations, ensuring its effectiveness and ethical application in various domains.

\nocite{langley00}

\bibliography{example_paper}
\bibliographystyle{icml2024}

%%%%%%%%%%%%%%%%%%%%%%%%%%%%%%%%%%%%%%%%%%%%%%%%%%%%%%%%%%%%%%%%%%%%%%%%%%%%%%%
%%%%%%%%%%%%%%%%%%%%%%%%%%%%%%%%%%%%%%%%%%%%%%%%%%%%%%%%%%%%%%%%%%%%%%%%%%%%%%%
% APPENDIX
%%%%%%%%%%%%%%%%%%%%%%%%%%%%%%%%%%%%%%%%%%%%%%%%%%%%%%%%%%%%%%%%%%%%%%%%%%%%%%%
%%%%%%%%%%%%%%%%%%%%%%%%%%%%%%%%%%%%%%%%%%%%%%%%%%%%%%%%%%%%%%%%%%%%%%%%%%%%%%%
\newpage
\appendix
\onecolumn
\section{Details of Dataset}

\subsection{Example of document-level data organization}
\label{app:Example of document-level data organization}
Instruction is the response generated by human experts or different LLMs. Output is the label of the data source as shown in Table \ref{tab:doc_lev_data_eg}.

% \vspace{-1.0cm}
\begin{table*}[h!]
\caption{ The example of the document-level data. }
\label{tab:doc_lev_data_eg}
\begin{tcolorbox}
{\bf Instruction:} \\
Categorize the texts into one of the two classes: human or AI. \\
{\bf Input:} \\ 
单间80多，如果住的天数多70多。里面有一个单独的卫生间，是隔出来的那种。其他的不是很清楚。另外德强那边有一个新开的巧克力时钟旅馆，听干净的，价格和相约谷差不多。你可以去看看。便宜点的还有恒久附近的，50多，但是没有单独的卫生间。都有电脑可以上网的。学校附近的小宾馆 旅社很多 学校里面也有非常多基本60-80左右 \\
(Translation: The single rooms are over 80, if you stay for more days it’s around 70. There’s a separate toilet inside, the kind that’s partitioned off. I’m not very clear about the others. By the way, there’s a newly opened Chocolate Clock Hotel over at Deqiang’s place, it’s said to be clean, and the price is about the same as Xiangyue Valley. You can go take a look. There are also cheaper ones near Hengjiu, over 50, but they don’t have a separate toilet. All of them have computers with internet access. There are many small hotels and guesthouses near the school, and there are also many inside the school, with prices ranging from about 60-80.) \\
{\bf Output:} Human \\
\end{tcolorbox}
\end{table*}

% \subsection{The prompt for polishing sentence}
% \label{app:The prompt for polishing sentence}
% To construct the AI-generated data, we adopt the ChatGPT to generate the sentences. We design a prompt to make the model understand the task and improve the performance. 
% The polishing prompt that we provide to ChatGPT is shown in Table \ref{tab:ai_sent_prompt}. 
% \begin{table}[ht]
% \caption{ The prompt for using ChatGPT to polish sentence-level data.}
% \label{tab:ai_sent_prompt}
% \begin{tcolorbox}
% {\bf Prompt} \\ 
% 请润色下述内容，不要做任何解释，直接输出润色结果： \\ 
% (Please polish the following content without any explanation, and output the polishing results directly:)
% \end{tcolorbox}
% \end{table}

\subsection{Example of sentence-level data organization}
\label{app:Example of sentence-level data organization}

%
%内容移动
%

The data source is the response written by human experts. If the sentence number of one piece of data is 7, we randomly select [L1, L3, L4, L7] as the sentence to be polished and hand it over to ChatGPT (3.5-turbo) for polishing. To construct the AI-generated data, we adopt the ChatGPT to generate the sentences. We design a prompt to make the model understand the task and improve the performance. 
The polishing prompt that we provide to ChatGPT is shown in Table \ref{tab:ai_sent_prompt}. 

\begin{table}[h!]
\caption{ The prompt for using ChatGPT to polish sentence-level data.}
\label{tab:ai_sent_prompt}
\begin{tcolorbox}
{\bf Prompt} \\ 
请润色下述内容，不要做任何解释，直接输出润色结果： \\ 
(Translation: Please polish the following content without any explanation, and output the polishing results directly:)
\end{tcolorbox}
\end{table}

% Finally, the OOD data be constructed as shown in Appendix \ref{app:Example of sentence-level data organization}.

% by using the prompt as shown in Table \ref{tab:1}. 

After obtaining the polished [P1, P3, P4, P7], splice them back together to form the paragraphs [P1, L2, P3, P4, L5, L6, P7] that blend AI and Human. Using the same method, we sampled 1,504 samples from M4 for sentence-level data construction. Ultimately, we used the HC3 sentence-level data as the training set and M4 as the test set. An example is shown in Table \ref{tab:sent_lev_data_eg}.
For sentence-level, the size of the train set is 5,589 and the size of the test set is 1,504.
% The details of the sentence-level dataset can be found in Table \ref{tab:sent_level_data_size}.

%
%内容移动
%

\begin{table*}[h!]
\caption{ The example of the sentence-level data. }
\label{tab:sent_lev_data_eg}
\begin{tcolorbox}
{\bf Instruction:} \\
Categorize the texts into one of the 2 classes: human or AI. \\
{\bf Input:} \\ 
"单间80多，如果住的天数多70多。里面有一个单独的卫生间，是隔出来的那种。其他的不是很清楚。另外德强那边有一个新开的巧克力时钟旅馆，听干净的，价格和相约谷差不多。你可以去看看。便宜点的还有恒久附近的，50多，但是没有单独的卫生间。都有电脑可以上网的。学校附近的小宾馆 旅社很多 学校里面也有非常多基本60-80左右 \\
(Translation: The single rooms are over 80, if you stay for more days it’s around 70. There’s a separate toilet inside, the kind that’s partitioned off. I’m not very clear about the others. By the way, there’s a newly opened Chocolate Clock Hotel over at Deqiang’s place, it’s said to be clean, and the price is about the same as Xiangyue Valley. You can go take a look. There are also cheaper ones near Hengjiu, over 50, but they don’t have a separate toilet. All of them have computers with internet access. There are many small hotels and guesthouses near the school, and there are also many inside the school, with prices ranging from about 60-80.)" \\
{\bf Output:} \textless HUMAN \textgreater单间80多，\textless /HUMAN \textgreater \textless HUMAN \textgreater如果住的天数多70多。\textless /HUMAN \textgreater \textless AI \textgreater里面有一个独立的卫生间，\textless /AI \textgreater \textless AI \textgreater是隔出来的那种。\textless /AI \textgreater \textless AI \textgreater其他方面有些不太清楚。\textless /AI \textgreater \textless AI \textgreater另外，\textless /AI \textgreater\textless AI \textgreater在德强那里有一家新开的巧克力时光旅馆，\textless /AI \textgreater \textless AI \textgreater听说环境很干净，\textless /AI \textgreater \textless AI \textgreater价格和相约谷差不多。\textless /AI \textgreater \textless AI \textgreater你可以去看看。\textless /AI \textgreater \textless /AI \textgreater在恒久附近有更便宜的房间，\textless /AI \textgreater \textless AI \textgreater大约50多元，\textless /AI \textgreater \textless AI \textgreater但是没有独立的卫生间。\textless /AI \textgreater \textless HUMAN \textgreater都有电脑可以上网的。\textless /HUMAN \textgreater \textless HUMAN \textgreater学校附近的小宾馆 旅社很多 学校里面也有非常多基本60-80左右\textless /HUMAN \textgreater \\
\end{tcolorbox}
\end{table*}

\subsection{The prompt for constructing the OOD dataset}
\label{app:The prompt for constructing the OOD dataset}
The prompt that we give to ChatGPT as shown in Table \ref{tab:ood_data_prompt}.
\begin{table}[h!]
\caption{ The prompt for constructing the OOD dataset.}
\label{tab:ood_data_prompt}
\begin{tcolorbox}
{\bf Prompt} \\ 
你是一个新闻编辑，用户给你一个新闻类别，请在用户指定类别要求下写一个100字到700字的新闻稿。新闻类别包括：政治新闻\ 经济新闻\ 社会新闻\ 科技新闻\ 文化艺术新闻\ 娱乐新闻\ 环境新闻 ...\\ 
(Translation: You are a news editor, and the user provides you with a news category. Write a news article of 100 to 700 words based on the specified category. The news classes include Political News, Economic News, Social News, Technology News, Cultural and Arts News, Entertainment News, and Environmental News, ...)
\end{tcolorbox}
\end{table}

\subsection{Details of the document-level dataset}
\label{app:Details of the document-level dataset}
There are three datasets for the document-level. Firstly, we collect data from HC3 and construct the train set including 21,681 Human-generated data and 96,453 AI-generated data. Second, we collect data from M4 and construct the in-domain test set including 3000 Human-generated data and 26,750 AI-generated data. Finally, we collect data from News Broadcast and construct the out-of-domain test set including 2000 Human-generated data and 1915 AI-generated data.
The details of the training set, in-domain test set, and OOD test set can be counted as shown in Table \ref{tab:doc_data_size}.
% \usepackage{multirow}
\begin{table}[h!]
    \caption{Document-Level Training and Test Sets with Different Data Sizes from Various Sources.}
    \label{tab:doc_data_size}
    \centering
    \resizebox{0.6\textwidth}{!}{%
    \begin{tabular}{cccccc}
        \hline
            No. & Train Set / Test Set & Data Label & Data Source & Count & Total \\ \hline
            1 & \multirow{9}{*}{Train Set} & Human & Human & 21681 & 21681 \\ \cline{3-6} 
            2 &  & \multirow{8}{*}{AI} & ChatGPT & 17376 & \multirow{8}{*}{96453} \\
            3 &  &  & ChatGLM2-6b & 12850 &  \\
            4 &  &  & XVERSE-13b & 12833 &  \\
            5 &  &  & Qwen-14b & 12823 &  \\
            6 &  &  & GPT-4 & 12796 &  \\
            7 &  &  & BlueLM-7b & 12702 &  \\
            8 &  &  & Baichuan2-53b & 12659 &  \\
            9 &  &  & ERNIE-Bot-3.5 & 2414 &  \\ \hline
            10 & \multirow{10}{*}{In-Domain Test Set} & Human & Human & 3000 & 3000 \\ \cline{3-6} 
            11 &  & \multirow{9}{*}{AI} & ChatGPT & 3000 & \multirow{9}{*}{26750} \\
            12 &  &  & ChatGLM2-6B & 3000 &  \\
            13 &  &  & XVERSE-13b & 2998 &  \\
            14 &  &  & Qwen-14b & 2997 &  \\
            15 &  &  & GPT-4 & 2987 &  \\
            16 &  &  & BlueLM-7b & 2980 &  \\
            17 &  &  & Davinci003 & 2975 &  \\
            18 &  &  & ERNIE-Bot-3.5 & 2972 &  \\
            19 &  &  & Baichuan2-53b & 2841 &  \\ \hline
            20 & \multirow{2}{*}{Out-of-Domain Test Set} & Human & News & 2000 & 2000 \\ \cline{3-6} 
            21 &  & AI & ChatGPT & 1915 & 1915 \\ \hline
            22 & Total & - & - & - & 151799 \\ \hline
    \end{tabular}
    }
\end{table}

\subsection{Dataset Source Parse Analysis}
\label{app:Dataset Source Parse Analysis}
\begin{figure}[h!]
  \centering
  \includegraphics[width=0.65\linewidth]{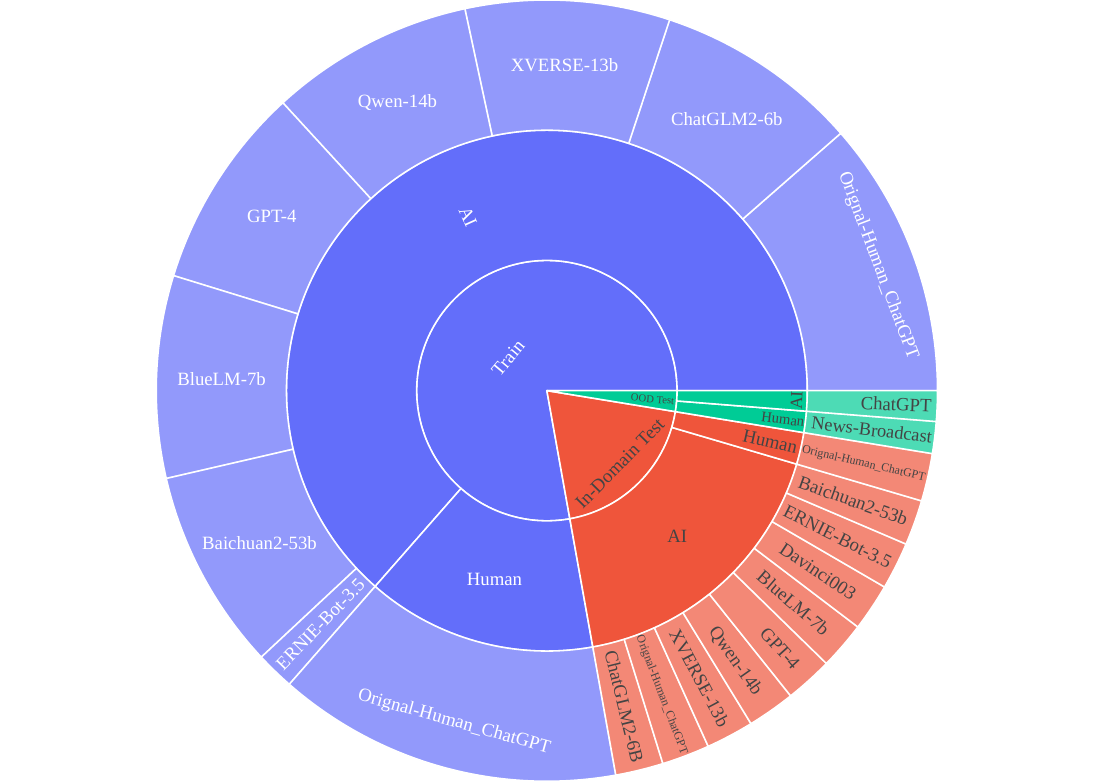}
  \caption{Dataset Source Parse Analysis. The source of the response data in the training set, in-domain test set, and OOD test set.}
  \label{fig:doc_data_pie}
\end{figure}

To explore and analyze our train set, in-domain test set, and out-of-domain test set, we do a dataset source parse analysis. The three datasets can be grouped by Haman and AI. Each grouped data can also be grouped again by its data source and can be plotted by the count of each part as shown in Figure \ref{fig:doc_data_pie}.

\subsection{Detailed analysis of the dataset}
\label{app:Detailed analysis of the dataset}
\begin{figure}[h!]
\vskip 0.2in
  \centering
  \includegraphics[width=0.8\linewidth]{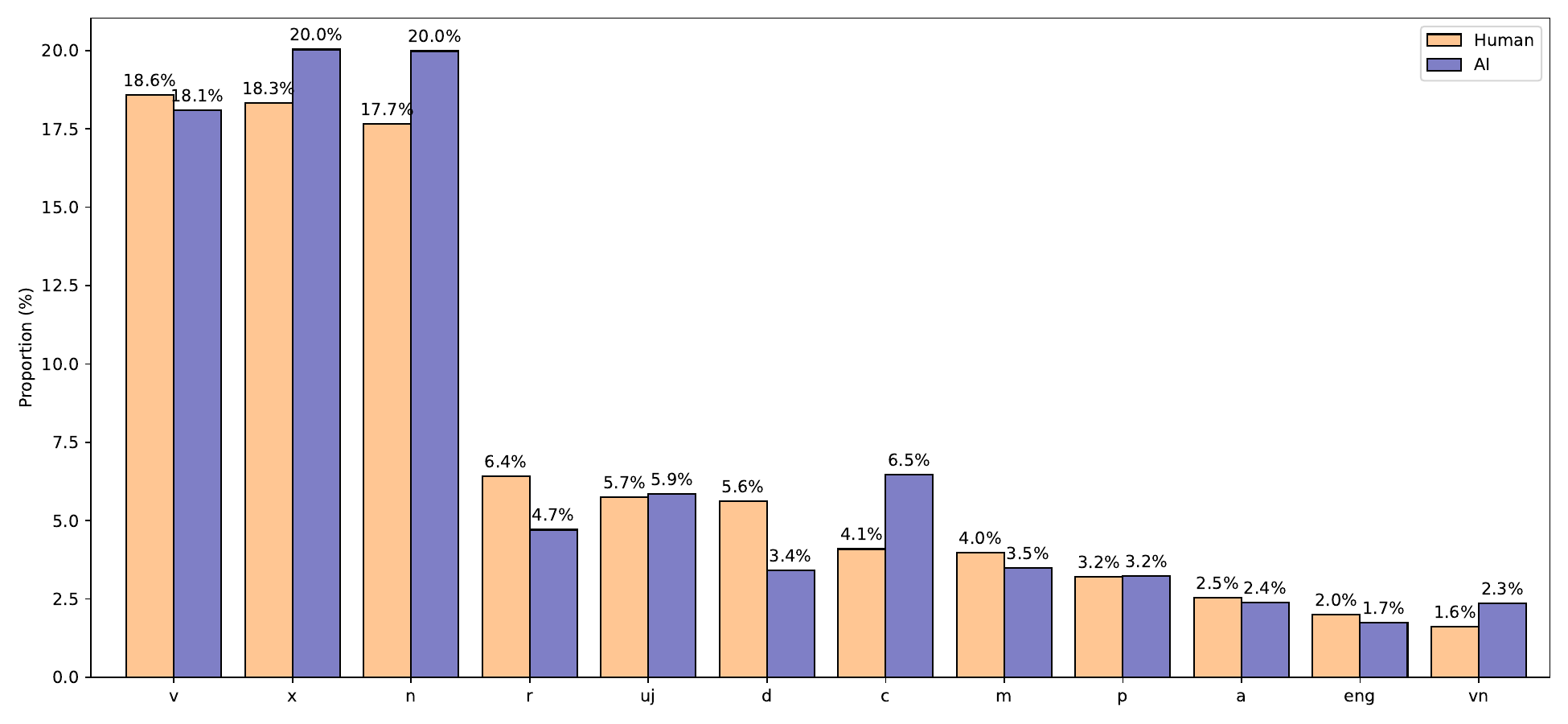}
  \caption{Part-of-Speech Comparison on Train Set.}
  \label{fig:part_of_speech_train_data}
  \vskip -0.2in
\end{figure}

\begin{figure}[h!]
\vskip 0.2in
  \centering
  \includegraphics[width=0.8\linewidth]{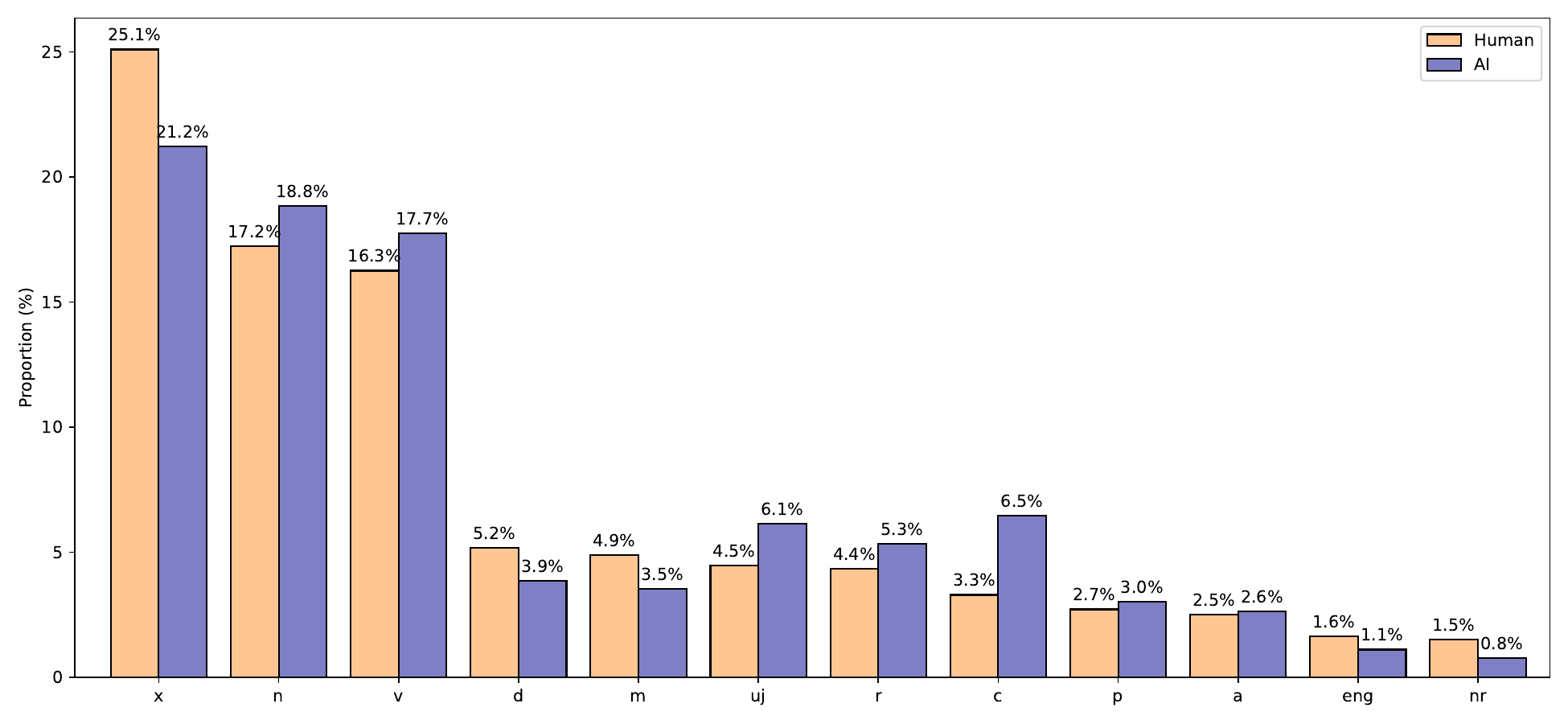}
  \caption{Part-of-Speech Comparison on Test Set.}
  \label{fig:part_of_speech_test_data}
  \vskip -0.2in
\end{figure}

In Figures~\ref{fig:part_of_speech_train_data} and~\ref{fig:part_of_speech_test_data}, we observe the distribution of part-of-speech tags in both training and test datasets, comparing human-written texts to those generated by AI. These figures highlight that while there is a general alignment in the linguistic structure of both sources, notable distinctions emerge in specific categories. In the training set (Figure~\ref{fig:part_of_speech_train_data}), human texts exhibit a marginally higher usage of verbs and nouns, whereas AI-generated texts have a slightly increased use of pronouns and adverbs. This trend is also evident in the test set (Figure ~\ref{fig:part_of_speech_test_data}), particularly with a marked increase in adverbs in AI texts, indicating a potential linguistic preference of the AI. Both figures corroborate that conjunctions and modal verbs are utilized with comparable frequency by humans and AI, suggesting a shared understanding of sentence construction and modality. The data encapsulated in these figures imply that while AI can closely emulate human part-of-speech patterns, distinct differences in usage can be pivotal in differentiating between human and AI-generated content.

\begin{figure}[h!]
\vskip 0.2in
    \centering
    \includegraphics[width=0.4\linewidth]{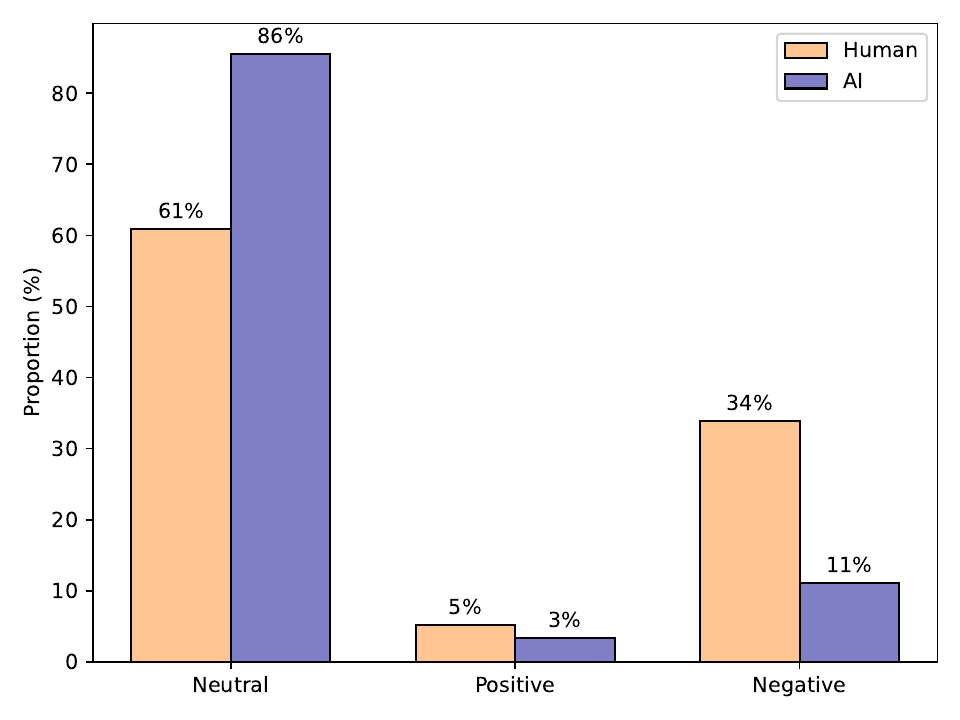}
    % \hfill
    \includegraphics[width=0.4\linewidth]{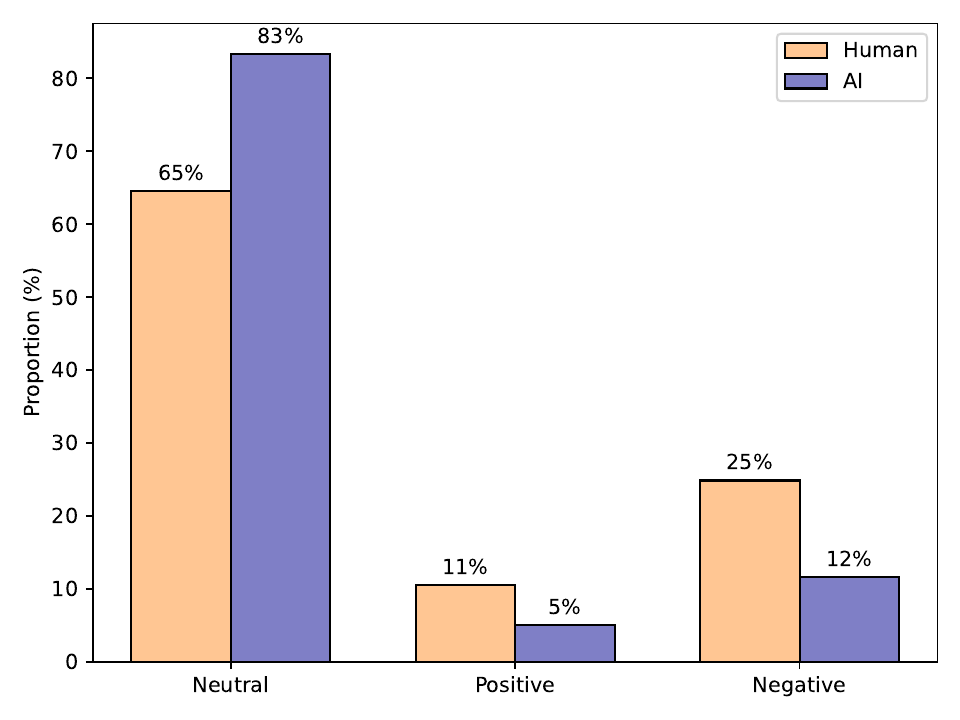}
    \caption{Sentiment Distribution: (Left) Distribution in the Training Set, (Right) Distribution in the Test Set.}
    \label{fig:sent_dist}
    \vskip -0.2in
\end{figure}

In Figure~\ref{fig:sent_dist}, the sentiment distribution across the training and test datasets for human-written and AI-generated texts is depicted. The bar charts compare the proportion of neutral, positive, and negative sentiments expressed in both datasets, with orange bars representing human-produced content and purple bars for AI-generated material. In the training set, a substantial majority of AI-generated texts are classified as neutral (86\%), while human texts show a slightly lower neutral sentiment proportion (61\%). Conversely, human texts exhibit a significantly higher inclination towards negative sentiments (34\%) compared to AI (11\%), with positive sentiments being minimal in both but slightly higher in AI (5\% compared to 3\% in human texts). A similar pattern is observable in the test set, where AI texts are predominantly neutral (83\%), but human texts are less so (65\%). The negative sentiment in human texts (25\%) is more than double that in AI texts (12\%), with positive sentiment remaining low for both. These charts suggest that AI-generated texts may tend toward neutral sentiment, while human authors express a broader emotional range, particularly negative sentiments. This pattern across both training and test sets highlights a key difference in the emotional tone between human and AI writing.

Our training set includes 119,475 neutral sentences which occupy 80.8\% of the dataset, 22,658 negative sentences which occupy 15.3\% of the dataset, and 5,751 positive sentences which occupy 3.9\% of the dataset.

\section{Details of the Experiment}

\subsection{Details of Tasks}
\label{app:Details of Tasks}
\textbf{In-domain Detection.} In-domain detection refers to the ability of the AI-generated text detector to accurately identify AI-generated text within the same domain or topic on which it was trained. To further evaluate the AI-generated text detector's performance in in-domain detection, we also experimented with different text lengths and evaluated its performance on fine-tuned LLMs with varying parameters. This allowed us to gain a better understanding of the AI-generated text detector's robustness and generalizability in in-domain text detection tasks.

\textbf{Out-of-domain Detection.} OOD detection refers to the ability of the AI-generated text detector to accurately identify AI-generated text from domains or topics different from those on which it was trained.

\textbf{Sentence-level Detection.} Sentence-level detection refers to the ability of an AI-generated text detector to accurately recognize AI-generated text at the sentence level (rather than at the document level). This is a more challenging task because the context provided by the entire document is not available, and the AI-generated text detector must rely solely on the content of a single sentence. 

\subsection{Details of Baselines}
\label{app:Details of Baselines}
To fully test the effectiveness of our proposed method, we compared it with AI-generated text detection methods based on statistical information and supervised learning methods. Additionally, we selected two advanced LLMs that have shown excellent performance in English.

\begin{itemize}
    \item Fast-DetectGPT~\cite{bao2023fast} is a method for zero-shot detection of AI-generated text. This method uses conditional probability curvature as an indicator and detects whether text is machine-generated by sampling and evaluating the differences in word selection probabilities. Compared to DetectGPT~\cite{mitchell2023detectgpt}, this method has increased detection speed by two orders of magnitude, while accuracy has improved by approximately 75\%.
    \item GLTR~\cite{gehrmann2019gltr} studied three types of features of an input text. Their major assumption is that to generate fluent and natural-looking text, most decoding strategies sample high probabilities tokens from the head of the distribution.
    \item Perplexity (PPL)~\cite{guo2023close} is a metric for evaluating the performance of language models. It measures the exponent of the negative average logarithmic likelihood of a given text under the language model. A lower PPL indicates that the language model is more confident in its predictions and is thus considered a better model. We use GPT-2 to calculate the PPL of human- and AI-generated content to distinguish who generated the text.
    \item MPU~\cite{textsmultiscale} proposes a multi-scale positive-unlabeled AI text detection method, which models AI text detection as a partial positive-unlabeled problem, utilizes length-based multi-scale PU loss, and introduces a text multi-scaling module. MPU significantly improves the detection performance of short texts and enhances long text detection. It has been implemented based on two methods: BERT and RoBERTa, referred to as BERT-MPU and RoBERTa-MPU, respectively.
    \item LLaMA-2~\cite{touvron2023llama} is a language model trained on approximately 2T tokens. It has demonstrated exceptional performance across multiple benchmark tests and has been widely used in LLM research. We adopt LLaMA-2-7B and LLaMA-2-13B as the base model for instruction tuning.
    \item Mistral-7B~\cite{jiang2023mistral} is a language model designed for superior performance and efficiency. It employs mechanisms such as grouped-query attention and sliding window attention to surpass other language models on various benchmarks.
\end{itemize}

In addition to the comparison methods mentioned above, we trained BERT~\cite{devlin2018bert} and RoBERTa~\cite{liu2019roberta} text classification models based on the same data for text detection. At the same time, we conducted Zero-shot text detection based on ChatGPT and GPT-4.

For classifying using ChatGPT and GPT-4 with Zero-shot, we conduct three predictions and take the average result. We adopted a method similar to that proposed by~\cite{holtzman2019curious} for open-text generation. Specifically, we used temperature sampling with a temperature is 0.7, top\_p is 1.0, and max\_tokens is 2048, while keeping other settings at their defaults. The prompts for ChatGPT and GPT-4 are as follows.

\begin{lstlisting}[breaklines]
Determine whether this passage is generated by AI or written by human. Do not respond 
with anything other than AI and Human. You are only allowed to answer AI or Human.
\end{lstlisting}

\subsection{Details of Metrics}
\label{app:Details of Metrics}
Commonly used concepts in evaluation metrics are expressed as follows:
\begin{itemize}
    \item True Positive (TP): the number of positive classes predicted to be positive classes.
    \item True Negative (TN): the number of negative classes predicted as negative classes.
    \item False Positive (FP): the number of negative classes predicted as positive classes, which is the number of detection errors.
    \item False Negative (FN): the number of positive classes predicted as negative classes, which is the number of missed detections.
\end{itemize}

Precision is defined as
\begin{equation}
\begin{aligned}
 Precision = \frac{TP}{TP+FP}
\end{aligned}
\end{equation}

Recall is defined as 
\begin{equation}
\begin{aligned}
Recall = \frac{TP}{TP+FN}
\end{aligned}
\end{equation}

Macro-F1 is defined as
\begin{equation}
\begin{aligned}
\text{Macro-F1} &= \frac{1}{N} \sum_{i=1}^{N} \frac{2 \times P_{i} \times R_{i}}{P_{i} + R_{i}} \\
% &= \frac{1}{N} \sum_{i=1}^{N} \frac{2 \times \text{TP}_{i}}{2 \times \text{TP}_{i} + \text{FP}_{i} + \text{FN}_{i}}
\end{aligned}
\end{equation}

$N$ represents the number of classes (In our task, it is two classes: Human and AI). $P_{i}$ and $R_{i}$ are the precision and recall for the $i$-th class, respectively. $\text{TP}_{i}$, $\text{FP}_{i}$, and $\text{FN}_{i}$ are the number of true positives, false positives, and false negatives for the $i$-th class, respectively. The Macro-F1 Score is calculated by taking the sum of the F1 Scores for each class and dividing by the total number of classes.

\subsection{The impact of text length on Fast-DetectGPT and RoBERTa}
\label{app:The impact of text length on Fast-DetectGPT and RoBERTa}

\begin{figure}[h!]
\vskip 0.1in
  \centering
  \includegraphics[width=0.65\linewidth]{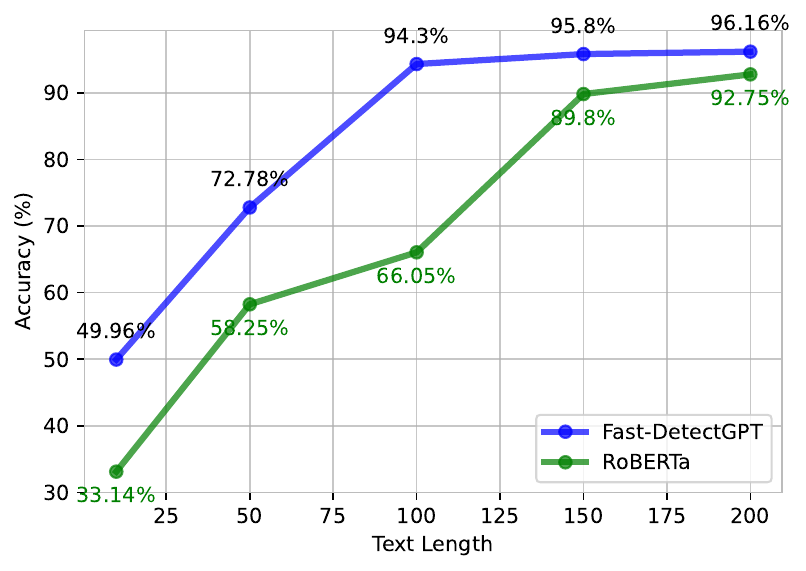}
  \caption{As the length of the text increases, the accuracy performance of Fast-DetectGPT and RoBERTa.}
  \label{fig:text_len}
  \vskip -0.1in
\end{figure}

We further investigated the impact of text length on Fast-DetectGPT and RoBERTa, as shown in Figure~\ref{fig:text_len}. We continued to sample texts of lengths 100, 150, and 200 from the in-domain dataset for detection. As the text length increased gradually, both Fast-DetectGPT (a statistical detector) and RoBERTa (a supervised detector) saw improvements in accuracy. After the text length exceeded 100 characters, the accuracy of Fast-DetectGPT rapidly rose to 94.3\%. When the text length exceeded 200 characters, the accuracy of RoBERTa rapidly increased to 83.8\%.

To evaluate the detection performance of different LLMs on content they have generated themselves, we fine-tuned the LLMs on responses generated by three different LLMs and human-written texts. The results are shown in the Figure~\ref{tab:diff_llm_diff_sour_acc}. A notable trend is that LLMs tend to perform best on text detection with texts they have generated. For instance, the ChatGLM2-6B model achieves the highest accuracy (99.91\%) on the dataset it generated, which is significantly higher than any other model tested against the same dataset. Similarly, the Qwen-14B model also has a high accuracy of 96.18\% on its generated dataset. However, an interesting anomaly arises with the BlueLM-7B model. The Qwen-7B model outperforms the BlueLM-7B on its own dataset, with an accuracy of 97.8\% compared to 97.1\% for the BlueLM-7B. While this could suggest a potential issue with the BlueLM-7B model's training, it is also worth noting that the difference is very small (only 0.7\%), which could fall within the margin of error.

\subsection{Are instruction-tuned LLMs better at detecting text they themselves have generated}
\label{app:Are instruction-tuned LLMs better at detecting text they themselves have generated}
\begin{table}[h!]
 \caption{Performance comparison of different LLMs based on different dataset sources. Bold text and blue background indicates the model with the best performance.}
  \centering
  \begin{tabular}{l|lc}
    \toprule
    % \cmidrule(r){1-3}
     Data Generation Source & Model & Accuracy \\ 
    \midrule
    ChatGLM2-6B & \cellcolor[RGB]{135,206,250} \textbf{ChatGLM2-6B} & \cellcolor[RGB]{135,206,250} \textbf{99.91\%} \\ 
    ChatGLM2-6B & XVERSE-7B & 94.60\% \\
    ChatGLM2-6B & Baichuan2-7B & 98.87\% \\
    ChatGLM2-6B & Qwen-7B & 96.63\% \\
    ChatGLM2-6B & Mistral-7B & 96.87\% \\
    ChatGLM2-6B & LLaMA-2-7B & 97.14\% \\
    \midrule
    Qwen-14B & \cellcolor[RGB]{135,206,250} \textbf{Qwen-14B} & \cellcolor[RGB]{135,206,250} \textbf{96.18\%} \\
    Qwen-14B & Baichuan2-13B & 95.92\% \\
    Qwen-14B & LLaMA-2-13B & 94.43\% \\
    Qwen-14B & XVERSE-13B & 91.19\% \\
    \midrule
    BlueLM-7B & BlueLM-7B & 97.10\% \\
    BlueLM-7B & XVERSE-7B & 92.51\% \\
    BlueLM-7B & Baichuan2-7B & 95.09\% \\
    BlueLM-7B & LLaMA-2-7B & 96.63\% \\
    BlueLM-7B & Mistral-7B & 94.44\% \\
    BlueLM-7B & \cellcolor[RGB]{135,206,250} \textbf{Qwen-7B} & \cellcolor[RGB]{135,206,250} \textbf{97.80\%} \\
    \bottomrule
  \end{tabular}
  \label{tab:diff_llm_diff_sour_acc}
\end{table}

\subsection{The impact of text generated by LLMs of different scales on the accuracy of text detection}
\label{app:The impact of text generated by LLMs of different scales on the accuracy of text detection}

We used the LLM-Detector to perform text detection on texts generated by LLMs of different parameter sizes. We found that the texts produced by LLMs of varying scales had no significant impact on the accuracy of text detection by LLM-Detector, indicating that detectors trained on LLMs demonstrate better robustness and generalization, as shown in Figure~\ref{tab:perf}. Specifically, the three differently sized detectors—Small, Medium, and Large showed a small range of fluctuation in detection accuracy for texts generated by LLMs of different scales, with the gap between the highest and lowest accuracy not exceeding 5\%.

\begin{table*}[h!]
\vskip 0.2in
 \caption{Performance comparison of different LLMs based on different dataset sources. The darker the color, the better the performance.}
  \centering
  \resizebox{\linewidth}{!}{
  \begin{tabular}{l|cc|cc|c|ccc}
    \toprule
     & \multicolumn{8}{c}{\cellcolor[gray]{0.90}\textbf{Gradually increasing model size range $\rightarrow$}} \\
     Model & ChatGLM2-6B & BlueLM-7B & XVERSE-13B & Qwen-14B & Baichuan2-53B & ERNIE-Bot & ChatGPT & GPT-4 \\
    \midrule
    LLM-Detector-Small & \cellcolor[RGB]{50,205,50}98.48\% & \cellcolor[RGB]{50,205,50}98.85\% & \cellcolor[RGB]{144,238,144}97.54\% & \cellcolor[RGB]{152,251,152}95.60\% & \cellcolor[RGB]{50,205,50}97.20\% & \cellcolor[RGB]{144,238,144}96.79\% & \cellcolor[RGB]{144,238,144}96.51\% & \cellcolor[RGB]{50,205,50}97.82\% \\
    LLM-Detector-Medium & \cellcolor[RGB]{144,238,144}95.41\% & \cellcolor[RGB]{152,251,152}95.89\% & \cellcolor[RGB]{152,251,152}95.84\% & \cellcolor[RGB]{50,205,50}99.33\% & \cellcolor[RGB]{152,251,152}96.83\% & \cellcolor[RGB]{152,251,152}96.49\% & \cellcolor[RGB]{152,251,152}95.81\% & \cellcolor[RGB]{152,251,152}94.48\% \\
    LLM-Detector-Large & \cellcolor[RGB]{152,251,152}93.28\% & \cellcolor[RGB]{144,238,144}98.11\% & \cellcolor[RGB]{50,205,50}99.44\% & \cellcolor[RGB]{144,238,144}98.15\% & \cellcolor[RGB]{144,238,144}96.99\% & \cellcolor[RGB]{50,205,50}99.23\% & \cellcolor[RGB]{50,205,50}99.64\% & \cellcolor[RGB]{144,238,144}95.25\% \\
    \bottomrule
  \end{tabular}}
  \label{tab:perf}
  \vskip -0.2in
\end{table*}

%%%%%%%%%%%%%%%%%%%%%%%%%%%%%%%%%%%%%%%%%%%%%%%%%%%%%%%%%%%%%%%%%%%%%%%%%%%%%%%
%%%%%%%%%%%%%%%%%%%%%%%%%%%%%%%%%%%%%%%%%%%%%%%%%%%%%%%%%%%%%%%%%%%%%%%%%%%%%%%

\end{CJK*}

\end{document}